\documentclass[11pt,a4paper]{article}
\usepackage[hyperref]{acl2020}
\usepackage{times}
\usepackage{latexsym}
\usepackage{amsfonts}
\usepackage{multirow}
\usepackage{amssymb}
\usepackage{subfigure}
\usepackage{graphicx}
\usepackage{arydshln}
\usepackage{amsmath}

\usepackage{microtype}

\aclfinalcopy 
\def\aclpaperid{2516} 

\title{How Can BERT Help Lexical Semantics Tasks?}

\author{Yile Wang, Leyang Cui, Yue Zhang \\
  Westlake University \\
  \texttt{\{wangyile, cuileyang\}@westlake.edu.cn} \\
  \texttt{yue.zhang@wias.org.cn}}

\date{}
\begin{document}
\maketitle
\begin{abstract}
Contextualized embeddings such as BERT can serve as strong input representations to NLP tasks, outperforming their static embeddings counterparts such as skip-gram, CBOW and GloVe. However, such embeddings are dynamic, calculated according to a sentence-level context, which limits their use in lexical semantics tasks. We address this issue by making use of dynamic embeddings as word representations in training static embeddings, thereby leveraging their strong representation power for disambiguating context information. Results show that this method leads to improvements over traditional static embeddings on a range of lexical semantics tasks, obtaining the best reported results on seven datasets.
\end{abstract}

\section{Introduction}
Word embedding~\cite{Bengio03} is a fundamental task in natural language processing, which investigates the representation of words in dense low-dimensional vector spaces. Seminal methods~\cite{miko2013a,glove,fasttext} are built based on the distributional hypothesis~\cite{harris1954}. In particular, the skip-gram model~\cite{miko2013a} trains a word embedding lookup table from large raw sentences by using a center word to predict its context words in a window size. Data likelihood is modeled based on cosine similarities between embeddings.

Word embeddings are useful in two broad aspects. First, they can be used directly to solve lexical semantics tasks, such as word similarity and analogy. For example, \citet{miko2013a} shows that the analogy between  $\langle${\it king}, {\it queen}$\rangle$ and $\langle${\it man}, {\it woman}$\rangle$ can be reflected by algebraic operations between the relevant word vectors. This characteristic of word embeddings is of large interest to computational linguistic research~\cite{we-lin1,we-lin2,we-lin3}. Second, word embeddings can be used as input representations to downstream tasks, addressing sparsity issues of one-hot or indicator feature functions.

Recently, contextualized word representation such as ELMo~\cite{elmo}, GPT~\cite{GPT,GPT2}, BERT~\cite{bert}, XLNet~\cite{xlnet} and RoBERTa~\cite{roberta} has been shown a more effective input representation method compared to traditional word representation such as skip-gram~\cite{miko2013a} and GloVe~\cite{glove}, giving significantly improved results in a wide range of NLP tasks, including question answering~\cite{choi-qa}, reading comprehension~\cite{bertrc} and commonsense reasoning~\cite{bert-commonsense}. Such embddings differ from traditional embeddings mainly in their parameterization. In addition to a lookup table, a sequence encoding network such as RNN and SAN is also used for calculating word vectors given a sentence. As a result, the vector for the same word varies under different contexts. We thus call them {\it dynamic embeddings}. In contrast, traditional methods are referred to as {\it static embeddings}.

One limitation of dynamic embeddings, as compared to static embeddings, is that they cannot be used directly for the aforementioned lexical semantics tasks due to the need for sentential contextualization. We investigate how to address this issue. Intuitively, there are several cheap methods to obtain static embeddings from dynamic embedding models. For example, the contextualized vectors of a word can be averaged over a large corpus. Alternatively, the word vector parameters from the token embedding layer in a contextualized model can be used as static embeddings. Our experiments show that these simple methods do not necessarily outperform traditional static embedding methods.

We consider integrating dynamic embeddings into the training process of static word embeddings, thereby fully exploiting their strength for obtaining improved static embeddings, and consequently for improving lexical semantics tasks. In particular, we integrate BERT and skip-gram by using BERT to provide dynamic embeddings for center words during the training of a skip-gram model. Compared with the skip-gram, the advantage is at least two-fold. First, polysemous words are represented using BERT embeddings, thereby resolving word sense ambiguities~\cite{sense}. Second, syntactic and semantic information over the entire sentence is integrated into the center word representation \cite{analysis-bert-1, analysis-bert-2}, thereby providing richer features compared to a word window.

Experiments over a range of lexical semantics datasets show that our method outperforms the existing state-of-the-art methods for training static embeddings, demonstrating the advantage of leveraging dynamic embeddings to improve lexical semantics tasks. To our knowledge, we are the first to systematically integrate contextualized embeddings for improving word similarity and analogy results.

\section{Related Work}
\noindent \textbf{Static Word Embeddings.} Skip-gram (SG) and continuous-bag-of-words (CBOW) are two models based on distributed word-context pair information~\cite{miko2013a}. The former predicts the context words for a center word, while the latter predicts a center word using its context words. \citet{ssg} claims that not all the context are equal and considered word order in the skip-gram model. \citet{dan14} and \citet{deps} further inject syntactic information by building word embeddings from the dependency parse trees over texts. GloVe~\cite{glove} learns word embeddings by factorizing global word co-occurrence statistics. Our model follows the skip-gram framework. The main difference between our work and the above methods is that center words are represented dynamically, rather than statically. 

Our work is related to a line of work on sense embedding~\cite{wordsense3, wordsense1,wordsense4}. However, they require a pre-defined set of senses, and rely on external word sense disambiguation for training static sense embeddings. In contrast, we use dynamic embeddings to automatically and implicitly represent senses.

\noindent \textbf{Dynamic Word Embeddings.} Contextualized word representations have been shown useful for NLP tasks~\cite{choi-qa,bertrc,bert-commonsense}. ELMo~\cite{elmo} provides deep word representations generated from LSTM based language modeling,  GPT~\cite{GPT, GPT2} improves language model pre-training based on Transformer~\cite{transformer}, BERT~\cite{bert} investigates self-attention-network for deep bidirectional representations, XLNet~\cite{xlnet} takes a generalized autoregressive pretraining model based on Transformer-XL~\cite{transxl}. 

The above models are based on a language modeling objective. However, they do not model word co-occurrences directly, which has been shown important for distributed word embeddings. By integrating dynamic embeddings into the training of static embeddings, we make use of both contextualized representation and co-occurrence information for improving lexical semantics tasks. 

\noindent \textbf{Dynamically Calculating Context Vectors for Word Embeddings.} SynGCN~\cite{syngcn} use graph convolution network (GCN) to integrate syntactic context for learning context embeddings. Our work is similar in dynamically calculating word representations. The main difference is that, while their model uses dependency parse trees and graph convolution network for better incorporating syntactic and semantic information, we directly model the sequential context by using BERT contextualized representation trained over large data.

\begin{figure*}[t]
	\centering  
	\subfigure[Skip-gram model.]{
		\includegraphics[width=0.45\textwidth]{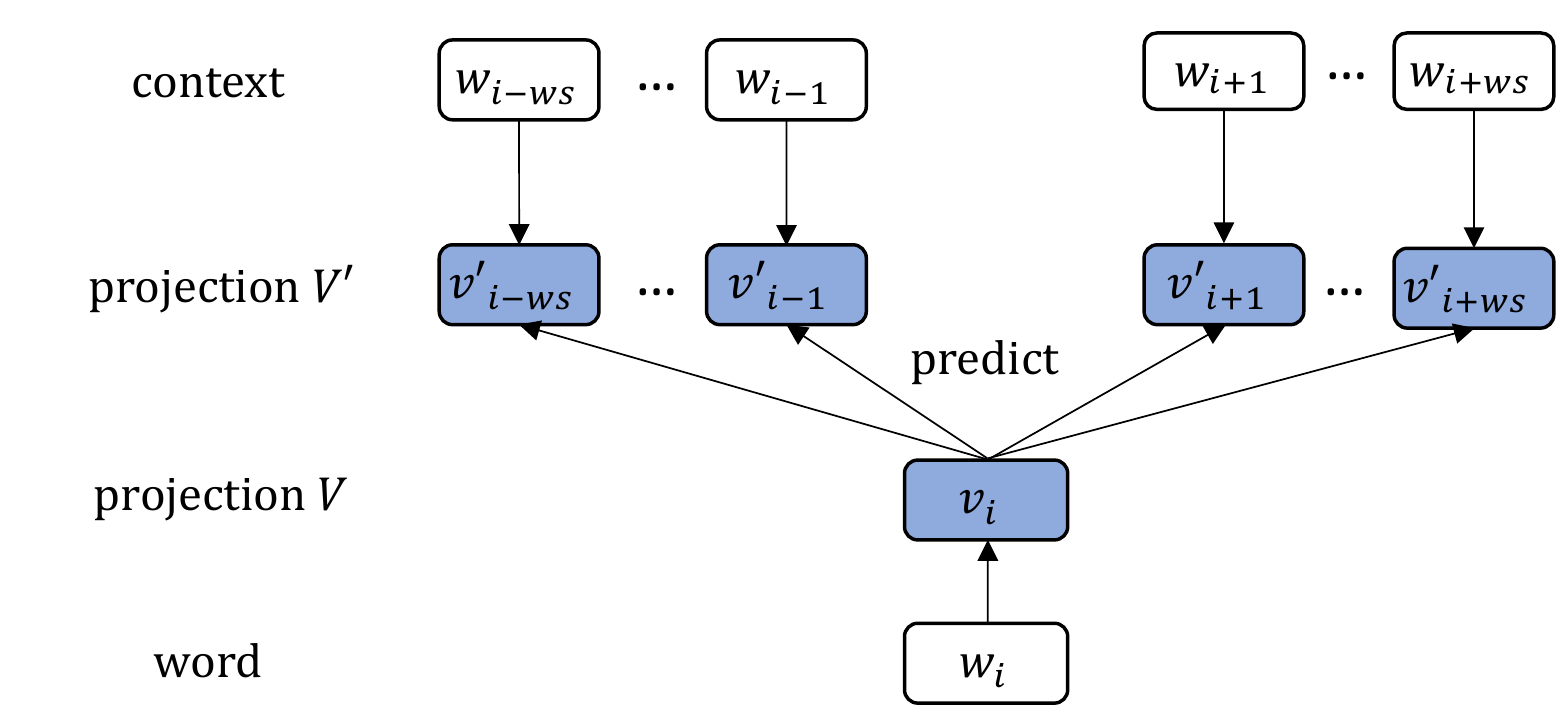}
		\label{figure:sg}
	}
	\subfigure[BERT model.]{
		\includegraphics[width=0.51\textwidth]{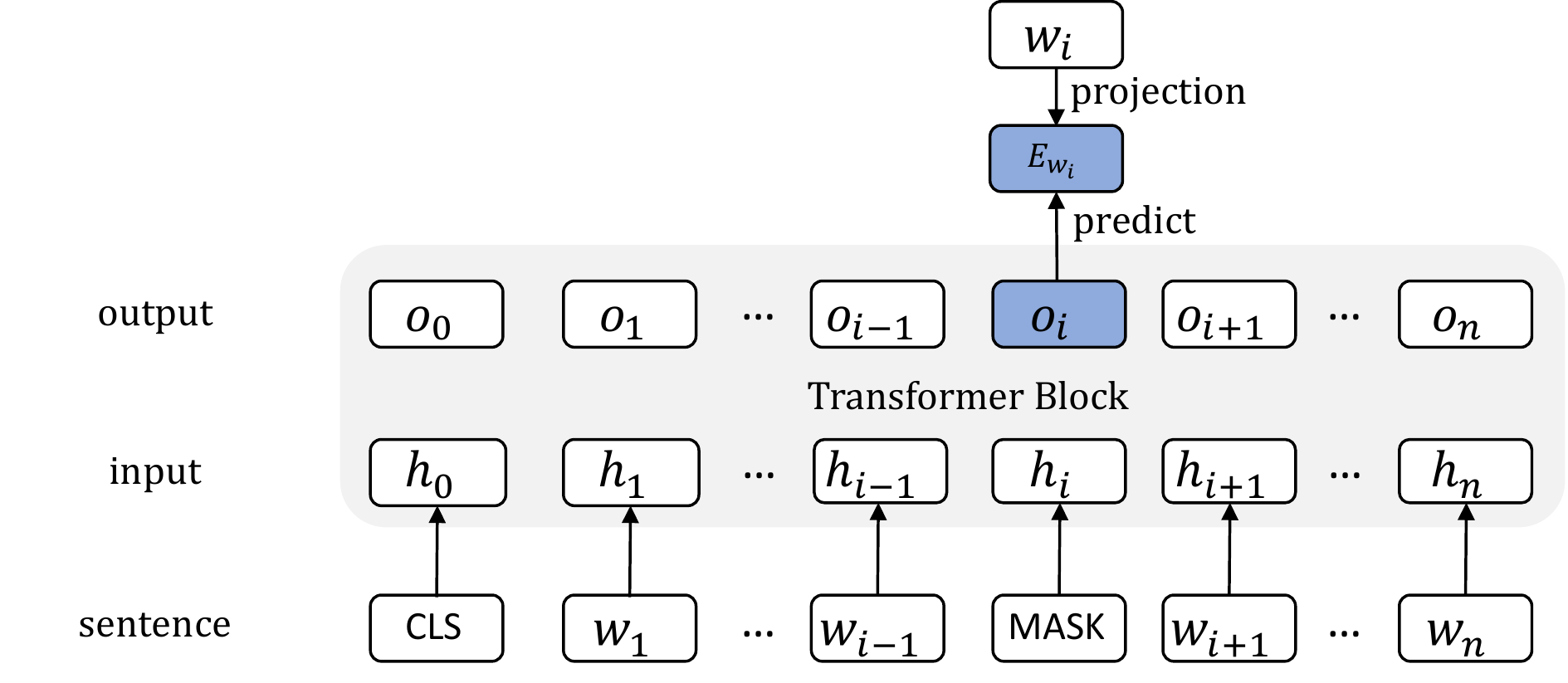}
		\label{figure:bert}
	}
	\subfigure[Our proposed model.]{
		\centering  
		\includegraphics[width=\textwidth]{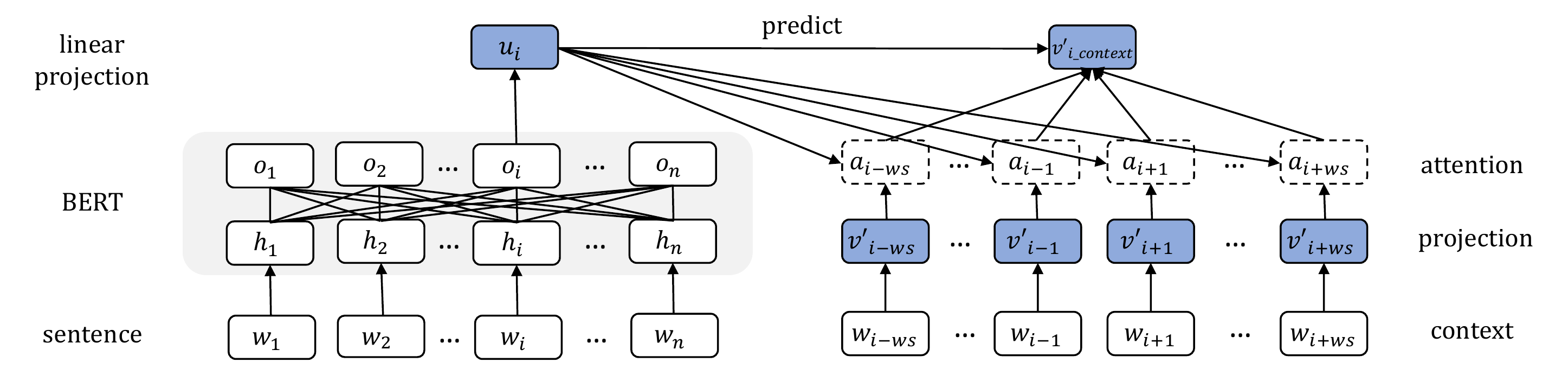}
		\label{figure:bertsg}
	}
	\centering
	\caption{Skip-gram, BERT and our proposed model. The blue blocks denote the representation of words.}
\end{figure*}

\section{Background}
We take skip-gram~\cite{miko2013a} as our base model for static word embeddings. BERT~\cite{bert} is used as the dynamic embeddings to replace the center word embeddings in our model.

\noindent \textbf{Skip-Gram.} Given a sentence $s$ with words $w_1, w_2, ..., w_{n} (w_i \in D)$, we model each word $w_i$ by using its context words $w_{i-ws}$, ..., $w_{i-1}$, $w_{i+1}$, ..., $w_{i+ws}$. The center word and context words are projected into two types of embeddings $v_{i}$ and $v'_{{i+j}} ({1}\le{\vert{j}\vert}\le{ws}$), respectively, as shown in Figure~\ref{figure:sg}. Given a training corpus with $N$ sentences $C = \{\textit{$s_c = w_{1}, w_{2}, ..., w_{n_c}$}\}|_{c=1}^{N}$,  the training objective is to minimize:
\begin{equation}
L_{\it{SG}} = -\sum_{c=1}^{N}\sum_{i=1}^{n_c}\sum_{{1}\le{\vert{j}\vert}\le{ws}} {\log f(v'_{i+j}, v_i)}
\label{eq:sg1}
\end{equation}
herein $f(v'_{i+j}, v_i) = p(w_{i+j}|w_i)$ represents the concurrence probability of word $w_{i+j}$ given the word $w_i$, which is estimated by:
\begin{equation}
p(w_{i+j}|w_i) = \frac{\exp({{v'_{i+j}}^\top{v_i})}}{\sum_{w_k \in D}\exp({{v'_{k}}^\top{v_i})}}
\label{eq:maxprob}
\end{equation}

During training, each word in the vocabulary uses the same embedding tables $V$ and $V'$ across sentences.


\noindent \textbf{BERT.} BERT consists of a multi-layer bidirectional Transformer~\cite{transformer} encoder. The masked language model (MLM) objective is to predict certain masked words through its contextualized representation, as shown in Figure~\ref{figure:bert}. 

Formally, given a sentence  $s = w_1, w_2, ..., w_{n}$, each $w_i$ is transformed into  input vector $h_i$ by summing up the static WordPiece~\cite{wordpiece} token embeddings \textit{$E_{w_i}$}, the segment embeddings  $\it{SE}_{w_i}$ and the position embeddings $\it{PE}_{w_i}$:
\begin{equation}
h_i = E_{w_i} + \it{SE}_{w_i} + \it{PE}_{w_i} 
\label{eq:e}
\end{equation}

The input vectors $H = \{h_1, ..., h_n\}$,  $H \in \mathbb{R}^{n\times{d}}$ are then transformed into queries $Q^m$, keys $K^m$, and values $V^m$, $\{Q^m, K^m, V^m\} \in \mathbb{R}^{n\times{d_k}}$:
\begin{equation}
Q^m, K^m, V^m = HW_Q^m, HW_K^m, HW_V^m
\label{eq:transformer1}
\end{equation}
where $\{W_Q^m, W_K^m, W_V^m\} \in \mathbb{R}^{d\times{d_k}}$ are trainable parameters, $m\in\{1,...,M\}$ represent the $m$-th attention head. $M$ parallel attention functions are applied to produce $M$ output states $\{O^1, ..., O^M\}$:
\begin{equation}
\begin{array}{l}
{A}^m = \it{softmax}(\frac{Q^{m\top}{K}^m}{\sqrt{d_k}}) \\
O^m = {A}^mV^m
\end{array}
\label{eq:transformer2}
\end{equation}

$A^m$ is the attention distribution for the $m$-th head and $\sqrt{d_k}$ is a scaling factor. Finally, each head for $O_i$ are concatenated to obtain the final output of word $w_i$:
\begin{equation}
o_i = [O^1_i, ..., O^M_i]
\label{eq:o}
\end{equation}


Given a corpus $\{\textit{$s_c = w_{1}, w_{2}, ..., w_{n_c}$}\}|_{c=1}^{N}$, the objective is to minimize the loss of predicting the randomly chosen masked word $w_{mask_i}$ by its output representation $o_{mask_i}$ in Eq.~\ref{eq:o}:
\begin{equation}
L_{\it{MLM}} = -\sum_{c=1}^{N}\sum_{i=1}^{n_c}{\log p(E_{w_{mask_i}}| o_{mask_i})}
\end{equation}
where $E$ is the token embedding table in Eq.~\ref{eq:e}, $p(E_{w_{mask_i}}| o_{mask_i})$ is calculated as with Eq.~\ref{eq:maxprob}:
\begin{equation}
p(E_{w_{mask_i}}| o_{mask_i}) = \frac{\exp({{E_{w_{mask_i}}}^\top{o_{mask_i}})}}{\sum_{w_{k} \in D}\exp({{E_{w_{k}}}^\top{o_{mask_i}})}}
\end{equation}

\section{The Proposed Approach}

Given a sentence $s = w_1, w_2, ..., w_{n}$, we model a center word $w_i$ and its context words $w_{i-ws}$, ..., $w_{i-1}$, $w_{i+1}$, ..., $w_{i+ws}$ as in the skip-gram model. To integrate dynamic embeddings, we use BERT to replace the center word embeddings $v_i$, so that each center word $w_i$ is represented in a sentential context. To this end, a center word $w_i$ is first transformed into $h_i$, which is the sum of the token embedding \textit{$E_{w_i}$} and the position embedding $\it{PE}_{w_i}$:
\begin{equation}
h_i = E_{w_i}  + \it{PE}_{w_i} 
\end{equation}


Then $h_1$, $h_2$, ..., $h_{n}$ are fed into a $L$-layer bidirectional Transformer block, as described in Eq.~\ref{eq:transformer1} and Eq.~\ref{eq:transformer2}. In particular, we use a pretrained BERT~\cite{bert} model to generate the output representations $o_i$, where numbers of layers $L=12$, attention heads $M=12$ and model size $d = 768$.

A linear projection layer is used for transforming the output $o_i \in \mathbb{R}^{d} $ to $u_{{i}}\in\mathbb{R}^{d_{emb}} $:
\begin{equation}
u_{{i}} = Uo_i
\end{equation}
where $U \in \mathbb{R}^{d_{emb}\times{d}}$ are model parameters.

To model co-occurrence between the center word $w_i$ and its context words $w_{i-ws}$, ..., $w_{i-1}$, $w_{i+1}$, ..., $w_{i+ws}$, we maximize the probability of the context words $w_{i+j} ( {1}\le{\vert{j}\vert}\le{ws})$ given the contextualized representation $u_i$ of the center word:
\begin{equation}
p(w_{i+j}|w_i) = \frac{\exp({{v'_{i+j}}^\top{u_i})}}{\sum_{w_k \in D}\exp({{v'_{k}}^\top{u_i})}}
\label{eq:maxprob2}
\end{equation}
similar to Eq.~\ref{eq:maxprob}, $v'_{k}$ is the context word embeddings for $w_{k}$ by using a static embedding table.

Note that our model is not a direct adaptation of the skip-gram model by replacing one embedding table. The original skip-gram algorithm uses the center word embedding table as the final output embeddings. However, to make the context words predictable and enable negative sampling from the vocabulary, we thus use BERT representation for the center word, and the context word embedding table as the final output static embeddings.

\textbf{Attention Aggregation.} Not all context words contribute equally to deciding the word representation. For example, predicting the stop words (e.g., \textit{``the''}, \textit{``a''}) is less informative than more meaningful words. One method to solve this problem is sub-sampling~\cite{miko2013b}. A word $w_i$ is discarded with a probability by:
\begin{equation}
P(w_i) = 1 - \sqrt{\frac{t}{f(w_i)}}
\label{eq:subsample}
\end{equation}
where $f(w_i)$ is the frequency of word $w_i$ in the training corpus and $t$ is a chosen threshold, typically around $10^{-5}$. 

\begin{figure*}[t]
	\centering
	\subfigure[]{
		\includegraphics[width=0.31\textwidth,trim= 0 0 0 35]{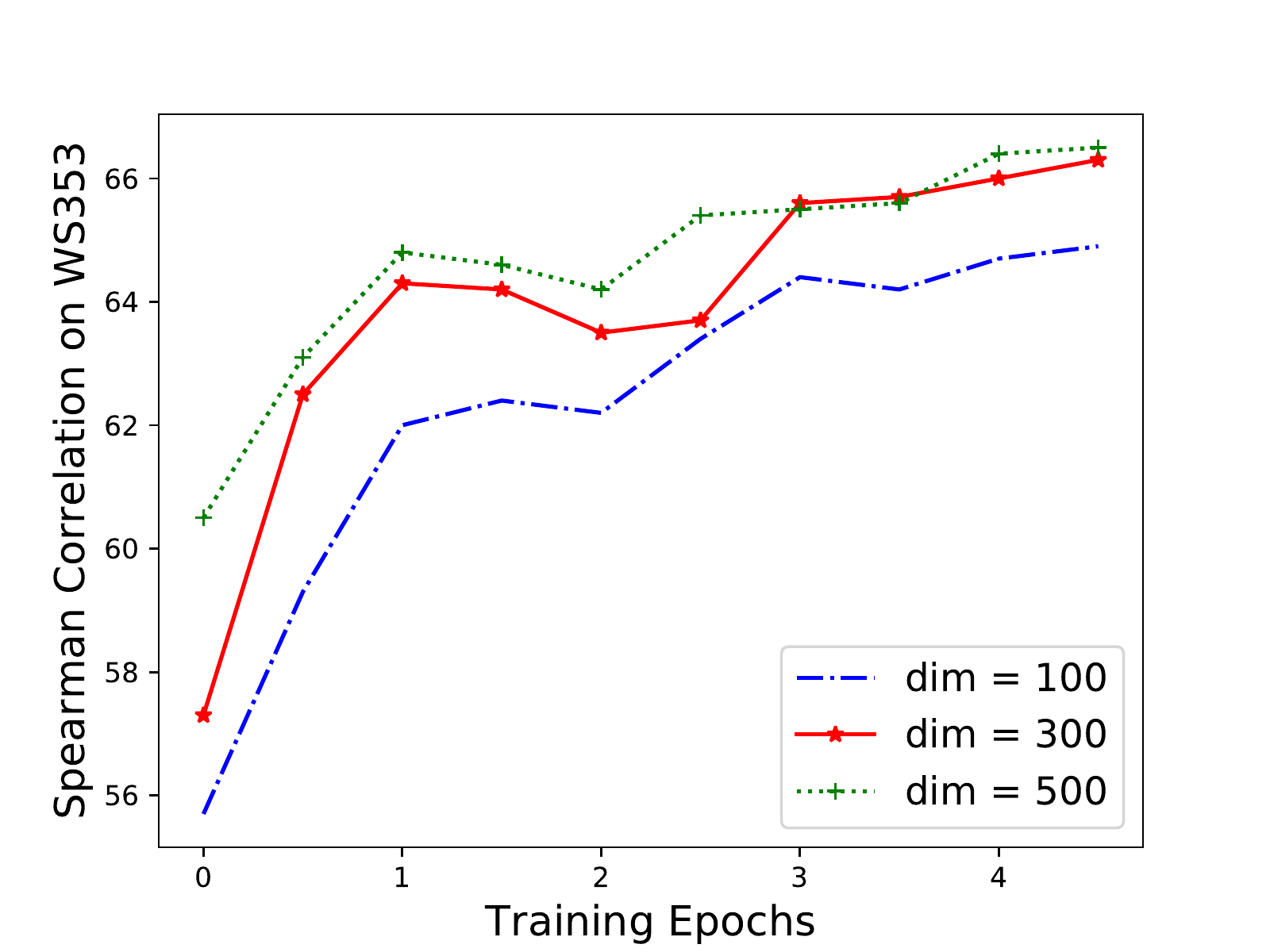}
		\label{figure:dev-dim}
	}  
	\subfigure[]{
		\includegraphics[width=0.31\textwidth,trim= 0 0 0 35]{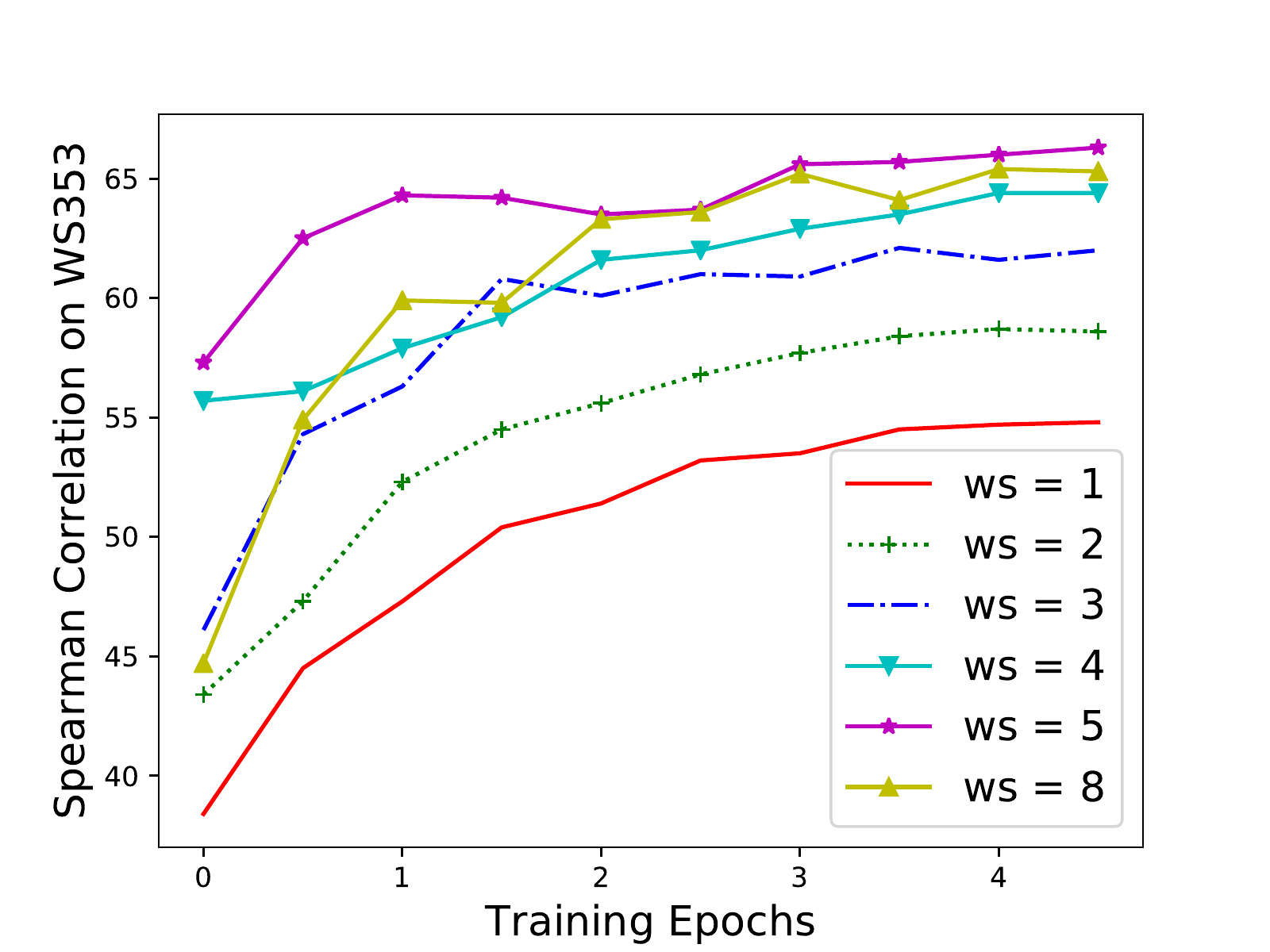}
		\label{figure:dev-ws}
	}
	\subfigure[]{
		\includegraphics[width=0.31\textwidth,trim= 0 0 0 35]{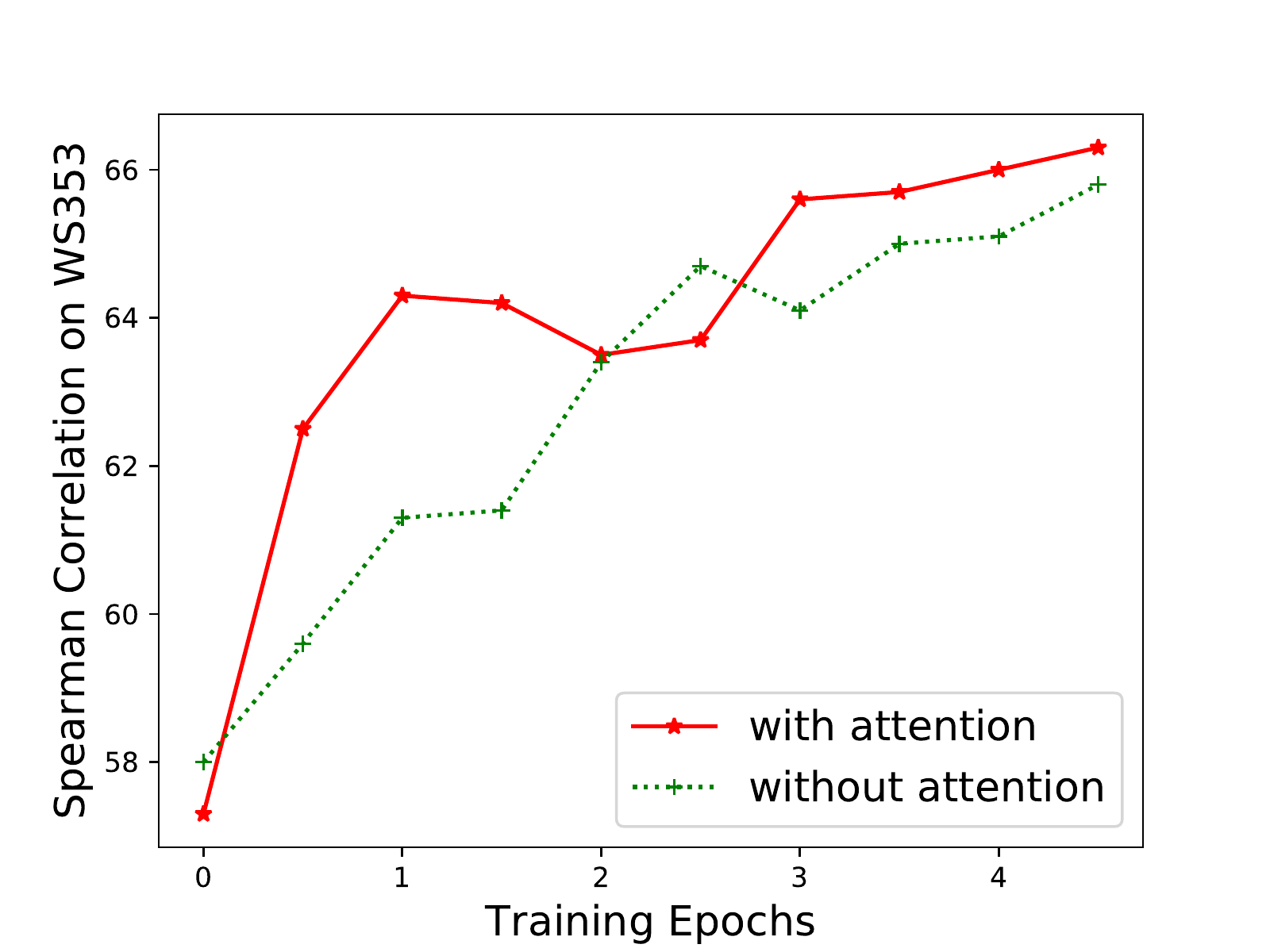}
		\label{figure:dev-att}
	}
	\centering
	\caption{Development experiments: (a) embedding dimension, (b) window size and (c) attention aggregation.}
\end{figure*}

Sub-sampling is used in the skip-gram model. However, it cannot be directly used in our method because contextualized representation can be undermined with words being removed from a sentence. We choose instead to select more indicative context words automatically while keeping the training sentence complete. Formally, we apply the attention mechanism to aggregate context words for each center word $w_j$ by using $u_{i}$ as the query vector and  $v'_j$ as the key vectors:
\begin{equation}
a_j = \textsc{ATT}(u_i, v'_j)
\label{eq:att}
\end{equation}
where $\textsc{ATT}(\cdot)$ denotes the dot-product attention operation~\cite{luong}.

The context embeddings are then combined using the corresponding attention coefficient:
\begin{equation}
v'_{i\_context} = \sum_{{1}\le{\vert{j}\vert}\le{ws}} {a_{i+j} v'_{i+j}}
\label{eq:att_context}
\end{equation}

\textbf{Training.} Given $\{\textit{$s_c = w_{1}, w_{2}, ..., w_{n_c}$}\}|_{c=1}^{N}$, the objective is to minimize a noise contrastive estimation loss function with negative sampling:
\begin{equation}
\begin{array}{l}
L = -\sum_{c=1}^{N}\sum_{i=1}^{n_c}\Big( {\log\sigma(v_{i\_context}^{'\top}{u_{i}})} \\
+ \sum_{m=1}^k\mathbb{E}_{w_{neg_m}\sim{P(w)}}[\log\sigma(-v_{neg_m}^{'\top}{u_{i}})]
\Big)
\end{array}
\end{equation}
where $\sigma$ is the sigmoid function, $w_{neg_m}$ denotes a negative sample, $k$ is the number of negative samples and $P(w)$ is the noise distribution set as the unigram distribution $U(w)$ raised to the 3/4 power (i.e., $P(w) = U(w)^{3/4}/Z$).

The final embeddings $v'$ are optimized through stochastic gradient descent.

\textbf{Testing.} The trained embeddings are tested for lexical semantics tasks in the following way. First, the similarity score between two words are calculated based on the cosine similarity between their embeddings:
\begin{equation}
\text{score}_{word} = \cos(x, y) = \frac{x^\top y}{\vert\vert x \vert\vert\cdot\vert\vert y\vert\vert}
\end{equation}

Second, the word analogy task investigates relations of the form ``$x$ is to $y$ as $x^*$ is to $y^*$'', where $y^*$ can be predicted given the word vectors of $x$, $y$, and $x^*$ by 3CosAdd~\cite{3cosadd}:
\begin{equation}
y^* = \mathop{\arg\max}_{y' \in V, y' \ne x^*, y, x} \cos((x^* + y - x), y')
\end{equation}

The relation similarity score between $x$ to $y$ and $x^*$ to $y^*$ is computed as:
\begin{equation}
\text{score}_{relation} = \cos((y - x), (y^* -x^*))
\end{equation}

\section{Experiments}
We compare the effectiveness of our method with both the skip-gram baselines and naive BERT methods for lexical semantics tasks. In addition, our methods are also compared with the state-of-the-art methods on standard benchmarks. 

\subsection{Experimental Settings}
\noindent \textbf{Datasets.} The Wikipedia dump\footnote{https://dumps.wikimedia.org/} corpus is used for training static embeddings, which consist of 57 million sentences with 1.1 billion tokens. Sentences with a length between 10 to 40 are selected, the final average length of sentences is 20.2. 

We perform word similarity tasks on the  WordSim-353~\cite{ws353}, SimLex-999~\cite{sim999}, Rare Word (RW)~\cite{rw} and MEN-3K~\cite{men} datasets, computing the Spearman's rank correlation between the word similarity score$_{word}$ and human judgments.

For word analogy, we compare the analogy prediction accuracy on the Google~\cite{miko2013a} datasets. We also compare the Spearman's rank correlation between relation similarity score$_{relation}$ and human judgments on the SemEval-2012~\cite{sem2012} dataset.

\noindent \textbf{Hyper-Parameters Settings.} The dimension of word embedding vectors $d_{emb}$ is 300; the window size for context words $\it{ws}$ is set as 5; the number of negative samples $k$ is 5; the initial learning rate for SGD is 0.08 and gradients are clipped at norm 5.


\begin{table*}[t]
	\centering
	\small
	\begin{tabular}{c|c|c|c|c|c|c|c|c|c}
		\hline 
		\multirow{2}*{\textbf{Types}}&\multirow{2}*{\textbf{Models}}&  \multicolumn{6}{c|}{\bf{Word Similarity}}& \multicolumn{2}{c}{\bf{Analogy}}\\
		\cline{3-10}
		& &\textbf{WS353}& \textbf{WS353S} & \textbf{WS353R}& \textbf{SimLex-999}& \textbf{RW}& \textbf{MEN}&  \textbf{Google}& \textbf{SemEval} \\
		\hline
		\multirow{5}*{Traditional}& SG &  61.0 &  68.9 &  53.7&  34.9&  34.5&  67.0&   43.5& 19.1 \\
		&CBOW &  62.7 &  70.7 &  53.9&  38.0&  30.0&  68.6&   58.4& 18.9 \\
		
		&GloVe &  54.2 &  64.3 &  50.2&  31.6&  29.9&  68.3&   45.3& 18.7 \\ 
		&FASTTEXT &  68.3 &  74.6 &  61.6&  38.2&  37.3&  74.8&    72.7& 19.5 \\
		&Deps &  60.6 &  73.1 &  46.8&  39.6&  33.0&  60.5&   36.0& 22.9 \\
		\cdashline{1-10} 
		\multirow{8}*{Dynamic Models}&ELMo$_{token}$ &  54.1 &  69.1 &  39.2&  41.7&  42.1&  57.7&   39.8& 19.3  \\
		
		&GPT2$_{token}$ &  65.5 & 71.5 &  55.7&  48.4&  31.6&  69.8&   33.1& 21.3 \\ 
		&BERT$_{token}$ &  57.8 &  67.3 &  42.5& 48.9&  29.5&  54.8&   31.7& 22.0 \\ 
		&XLNet$_{token}$ &  62.4 &  74.4 &  53.2&  48.1&  34.0&  66.3&   32.6& 22.2 \\ 
		
		&ELMo$_{avg}$ &  58.3 &  71.3 &  47.4&  43.6&  38.4& 65.5&  49.1& 21.2 \\
		&GPT2$_{avg}$ &  64.5 &  72.1 &  59.7&  46.9&  29.1&  68.6&   37.2& 21.9 \\
		&BERT$_{avg}$ &  59.4 &  67.0 &  49.9&  46.8&  30.8&  66.3&   59.4& 20.8 \\
		&XLNet$_{avg}$ &  64.9 &  72.3 &  58.0&  47.3&  27.7&  64.1&   30.8& 23.2 \\
		\cdashline{1-10} 
		\multirow{1}*{GCN + Static}&SynGCN &  60.9 &  73.2 &  45.7&  45.5&  33.7&  71.0&  50.7& \textbf{23.4} \\
		\cdashline{1-10} 
		\multirow{1}*{BERT + Static}&Ours &  \textbf{72.8} &  \textbf{75.3} &  \textbf{66.7}&  \textbf{49.4}&  \textbf{42.3}&  \textbf{76.2}&   \textbf{75.8}& 20.2 \\
		\hline
	\end{tabular}
	\caption{Main results on word similarity and analogy tasks. The ELMo, GPT2, BERT and XLNet models use 512, 768, 768 and 768 dimensional embeddings, respectively, while others use 300 dimensional vectors.}
	\label{table:ws_wa}
\end{table*}

\subsection{Development Experiments}
We select one million sentences from corpus for development experiments, investigating the effect of embedding dimension, context window size and attention aggregation.


\noindent \textbf{Embedding Dimension.} Figure~\ref{figure:dev-dim} shows the results for different word embedding dimension $d_{emb}$. The model with 100 dimensional embeddings gives a lower result, which is likely because the model will underfit with too few dimensions. The model with 500 dimensions gives similar final result compared with 300 dimensions, while having more parameters and taking more training and testing time. We thus select the dimension as 300, which is the same as most traditional models.

\noindent \textbf{Window Size.} The window size $\it{ws}$ decides how much context information we directly model. The results are shown on Figure~\ref{figure:dev-ws}. When $\it{ws}$ is 1, we only model the relationship between the center word and its two neighbor words. The performance is 54.8. As the window size increases, the model gives better results, showing the effectiveness of modeling more context information. However, when the window size is 8, the model costs twice as much training time but does not give further improvements. Therefore we set the window size to 5, which is the same as skip-gram.

\noindent \textbf{Attention Aggregation.} Figure~\ref{figure:dev-att} shows the  results of skip-gram and our model with or w/o attention aggregation. Our model stably outperforms skip-gram. Without attention aggregation, our model treats all context words equally. It gives slower convergence with a best development result of 65.5, lower than 66.3 with attention aggregation. This shows the effectiveness of differentiating context words~\cite{miko2013b}.

\subsection{Baselines}
\noindent$ \bullet\ \bf{SG,CBOW}$ are the skip-gram and continuous-bag-of-words models  by \citet{miko2013a}.

\noindent $\bullet\ \bf{GloVe}$ is a log-bilinear regression model which leverages global co-occurrence statistics of corpus~\cite{glove}.

\noindent $\bullet\ \bf{FASTTEXT}$ takes into account subword information that incorporates character n-grams into the skip-gram model~\cite{fasttext}.

\noindent $\bullet\ \bf{Deps}$ modifies the skip-gram model using the dependency parse tree to replace the sequential context~\cite{deps}. 

\noindent $\bullet\ \bf{BERT.}$ We investigate two simple ways to use BERT~\cite{bert} for lexical semantics tasks. The first method, called BERT$_{token}$, ignores the contextualized parameters and uses the mean pooled subword token embeddings from $E$ in Eq.~\ref{eq:e} as a set of static embeddings. The second method, called BERT$_{avg}$, takes the avarage of output $o_i$ in Eq.~\ref{eq:o} over training corpus.

\noindent $\bullet\ \bf{ELMo, GPT2\  and\  XLNet.}$ Similar to  BERT, we also investigate the token embeddings and the average of output representation from ELMo~\cite{elmo},  GPT2~\cite{GPT2} and XLNet~\cite{xlnet} models. The baselines are 
ELMo$_{token}$, ELMo$_{avg}$, GPT2$_{token}$, GPT2$_{avg}$, XLNet$_{token}$ and XLNet$_{avg}$, respectively.

\noindent $\bullet\  \bf{SynGCN.}$ Given a training sentence, ~\citet{syngcn} propose a GCN to dynamically calculate context word embeddings based on the syntax structure, using this dynamically calculated embeddings in training static embeddings.

The above baselines can be categorized into three classes, as shown in the first column in Table~\ref{table:ws_wa}. In particular, the first category of methods are traditional static embeddings, where word vectors come from a lookup table. In the second category, static embeddings from dynamic word embedding models are used. In the last category, dynamic and static embeddings are integrated in the sense that context or center words representation are dynamically calculated via GCN or a BERT model for each sentence, but the target embeddings are static.

\begin{table*}[t!]
	\centering
	\small
	\begin{tabular}{c|c|c|c|c|c|c}
		\hline 
		\textbf{Word Pairs} & \textbf{Human}& \textbf{SG ($\Delta$)}& \textbf{BERT$_{token}$ ($\Delta$)}& \textbf{BERT$_{avg}$ ($\Delta$)}& \textbf{SynGCN ($\Delta$)} & \textbf{Ours ($\Delta$)} \\
		\hline
		\textit{dividend}, \textit{payment} &  0.763 & 0.464 (-0.299)  &  0.347 (-0.416) &0.503 (-0.260)&0.431 (-0.332) &\textbf{0.566 (-0.197)}\\ 
		\textit{murder}, \textit{manslaughter} &  0.853 & 0.600 (-0.253) & 0.369 (-0.484) & 0.672 (-0.181)& 0.516 (-0.337)& \textbf{0.712 (-0.141)}\\ 
		\textit{shower}, \textit{thunderstorm}& 0.631 & 0.401 (-0.230) & 0.344 (-0.287)& 0.483 (-0.148)& 0.398 (-0.233)&\textbf{0.496 (-0.135)}\\
		\textit{board},	\textit{recommendation} & 0.447  & 0.259 (-0.188)  & 0.299 (-0.148)& 0.583 (+0.136)&0.092 (-0.355) &\textbf{0.342 (-0.105)}\\
		\textit{benchmark},	\textit{index} & 0.425  & 0.305 (-0.120) & 0.247 (-0.178)& 0.569 (+0.144)&0.255 (-0.170)& \textbf{0.435 (-0.010)}\\

		\hline
	\end{tabular}
	\caption{Word similarity comparison between human and models. The scores of human are normalized to (0,1).}
	\label{table:ana_ws}
\end{table*}

\begin{table*}[t!]
	\centering
	\small
	\begin{tabular}{c|c|c|c|c|c|c}
		\hline 
		\textbf{Types} & \textbf{Example}& \textbf{SG}& \textbf{BERT$_{token}$}& \textbf{BERT$_{avg}$}& \textbf{SynGCN} & \textbf{Ours} \\
		\hline
		capital-country & \textit{Berlin}  to \textit{Germany} is \textit{Ottawa}  to \textit{Canada}  & 59.7 & 17.2 & 45.3& 51.3& \textbf{86.7}\\ 
		city-state&  \textit{Phoenix}  to \textit{Arizona} is \textit{Dallas}  to \textit{Texas} & 39.2  & 16.2 & 36.2& 38.4& \textbf{70.8}\\
		nationality-adjective&  \textit{Austria}  to \textit{Austrian} is \textit{Spain}  to \textit{Spanish} & 67.3 &69.3 & 87.9& 40.1& \textbf{90.3}\\
		family& \textit{son}  to \textit{daughter} is \textit{uncle}  to \textit{aunt} & 63.6  & 41.5 & 76.6& 69.5& \textbf{86.7}\\
		comparative&  \textit{good}  to \textit{better} is \textit{easy}  to \textit{easier} & 53.4  & 55.2 & 80.4& 78.6& \textbf{91.7}\\
		superlative&  \textit{fast}  to \textit{fastest} is  \textit{bad}  to \textit{worst} & 23.8  & 41.6  & 58.0& 45.5& \textbf{85.9}\\
		plural&  \textit{dog}  to \textit{dogs} is \textit{mouse}  to \textit{mice} & 38.5  & 28.3&90.6 & 74.7& \textbf{92.2}\\
		\hline
	\end{tabular}
	\caption{Word analogy prediction accuracy on Google datasets according to different types of word pairs.}
	\label{table:ana_google}
\end{table*}

\subsection{Results}
Table~\ref{table:ws_wa} shows the main results on word similarity and analogy tasks. Our model gives the best performance on 7 out of 8 datasets, achieving the best results on all the word similarity datasets ($p$-value $<$ 0.01 using t-test). In particular, it outperforms the best performing baselines by a large margin on WS353, WS353R and Google datasets, obtaining 6.5\%, 8.3\%, and 4.2\% improvement, respectively.

Among the traditional word embedding baselines, the skip-gram and CBOW models give relatively similar results. The FASTTEXT model gives the best result for word similarity tasks by leveraging subword information. The syntax-based embeddings Deps outperforms other traditional embeddings on the SemEval-2012 dataset. The reason can be that the syntax-based embedding encodes functional similarity rather than topical similarity~\cite{dep16}, which is more suitable for the relation similarity tasks,  including relation classes such as ``part-whole" (e.g., $\langle$\textit{car}, \textit{engine}$\rangle$ is more similar to $\langle$\textit{hand}, \textit{finger}$\rangle$ than $\langle$\textit{bottle}, \textit{water}$\rangle$) and ``cause-purpose" (e.g., $\langle$\textit{anesthetic}, \textit{numbness}$\rangle$ is more similar to $\langle$\textit{joke}, \textit{laughter}$\rangle$ than $\langle$\textit{smile}, \textit{friendship}$\rangle$). 


With regard to dynamic embedding models, the static token embeddings (e.g., BERT$_{token}$) and the average of output representations (e.g., BERT$_{avg}$) perform relatively close on word similarity tasks, giving comparable results on some datasets such as WS353S and RW, and better than traditional models on the SimLex-999 and SemEval-2012 datasets. This shows the effectiveness of dynamic and sentential information. However, these methods do not fully exploit the word co-occurrence information, and thus still underperform static baselines on datasets such as WS353, WS353R and MEN. Our method outperforms these methods for using dynamic embeddings, showing the usefulness of integrating dynamic embeddings into static embedding training based on the distributional hypothesis.


\begin{figure*}[t]
	\centering  
	\subfigure[SG]{
	\begin{minipage}[t]{0.3\linewidth}
	    \centering  
    	\fbox{\includegraphics[width=0.88\textwidth,trim= 80 60 60 60]{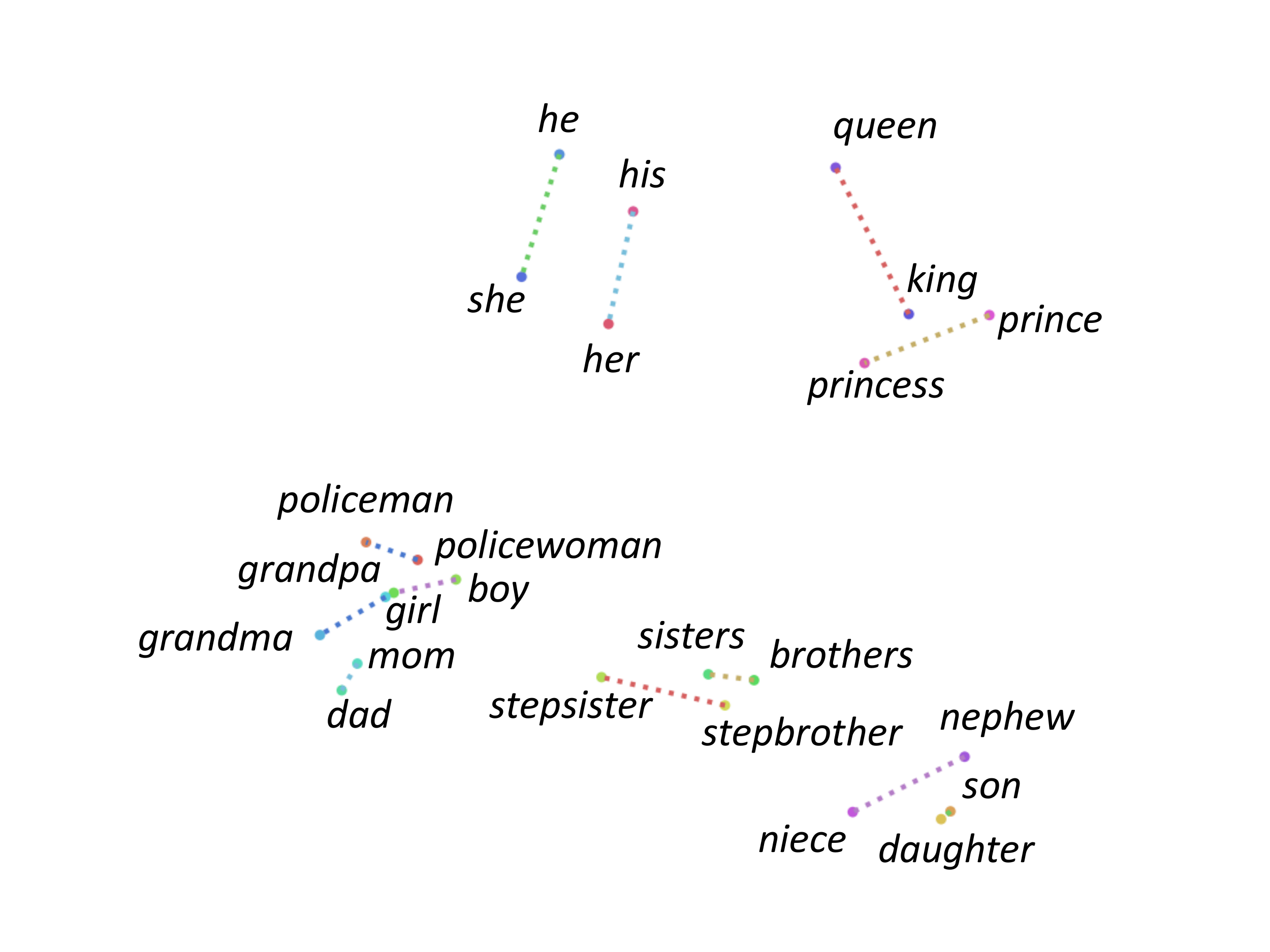}}
    \end{minipage}%
    }
    \subfigure[CBOW]{
	\begin{minipage}[t]{0.3\linewidth}
	    \centering  
    	\fbox{\includegraphics[width=0.88\textwidth,trim= 80 60 60 60]{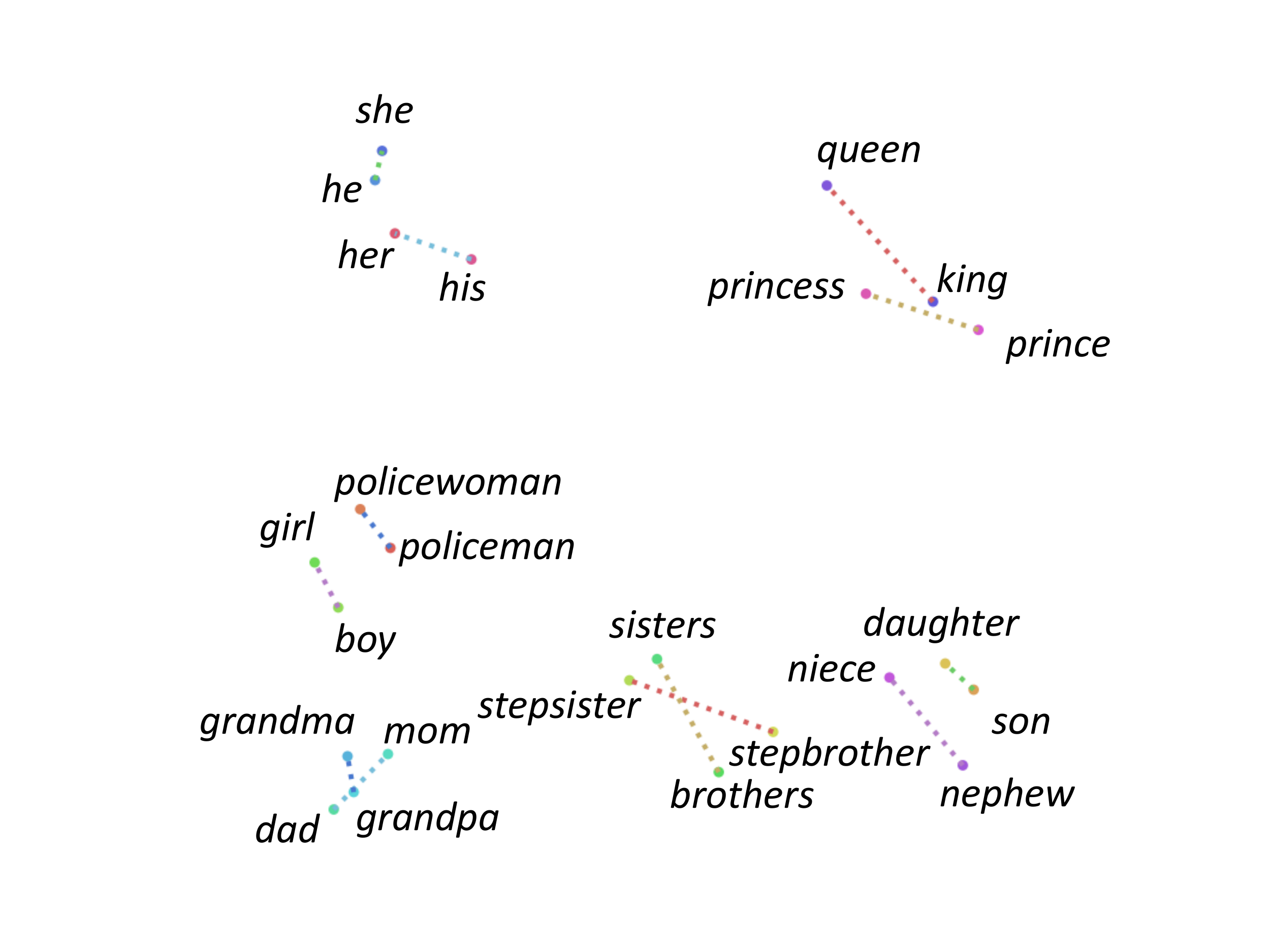}}
    \end{minipage}%
    }
    \subfigure[GloVe]{
	\begin{minipage}[t]{0.3\linewidth}
	    \centering 
    	\fbox{\includegraphics[width=0.88\textwidth,trim= 80 60 60 60]{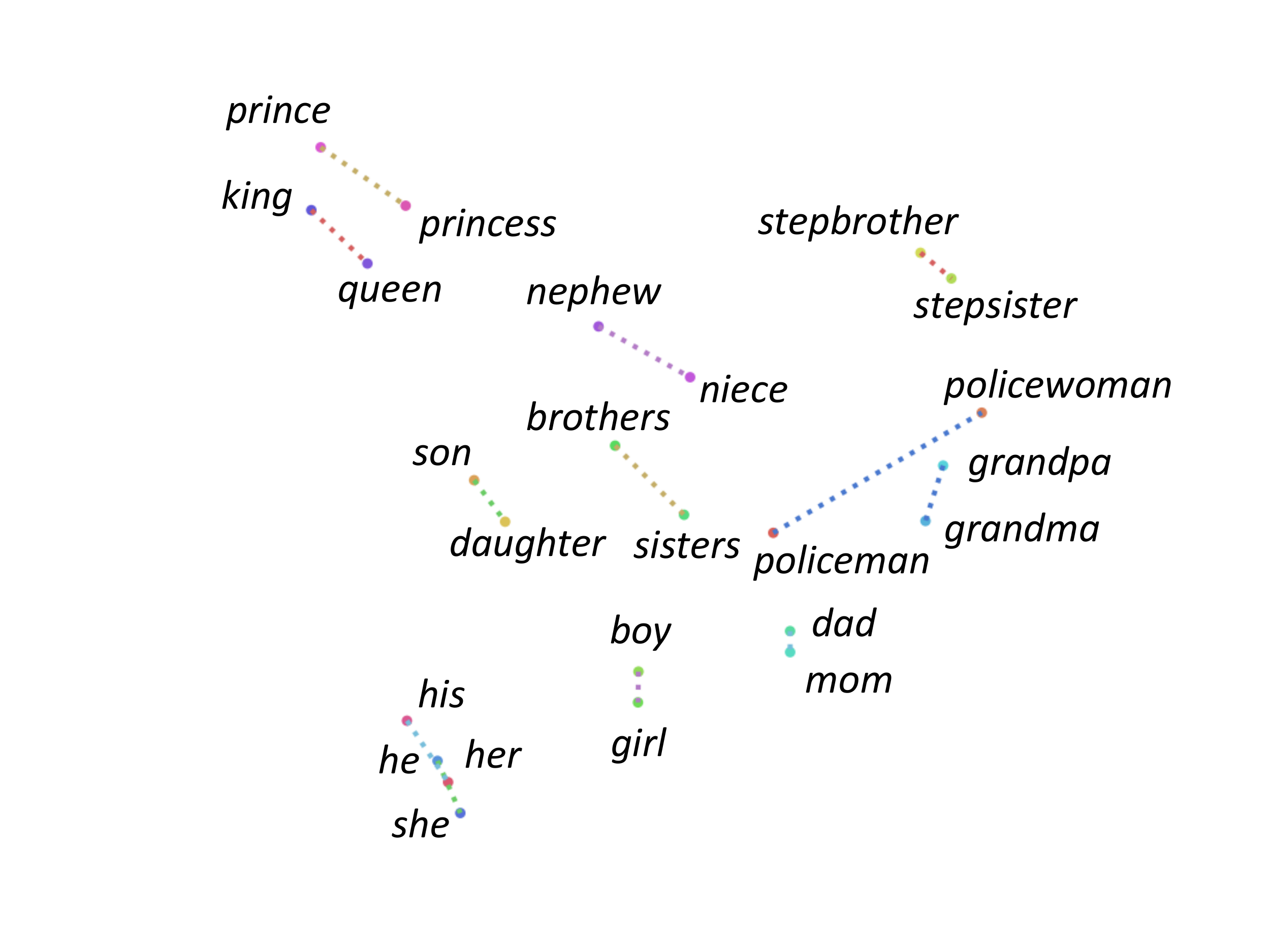}}
    \end{minipage}%
    }
    \subfigure[FASTTEXT]{
	\begin{minipage}[t]{0.3\linewidth}
	    \centering  
    	\fbox{\includegraphics[width=0.88\textwidth,trim= 80 60 60 60]{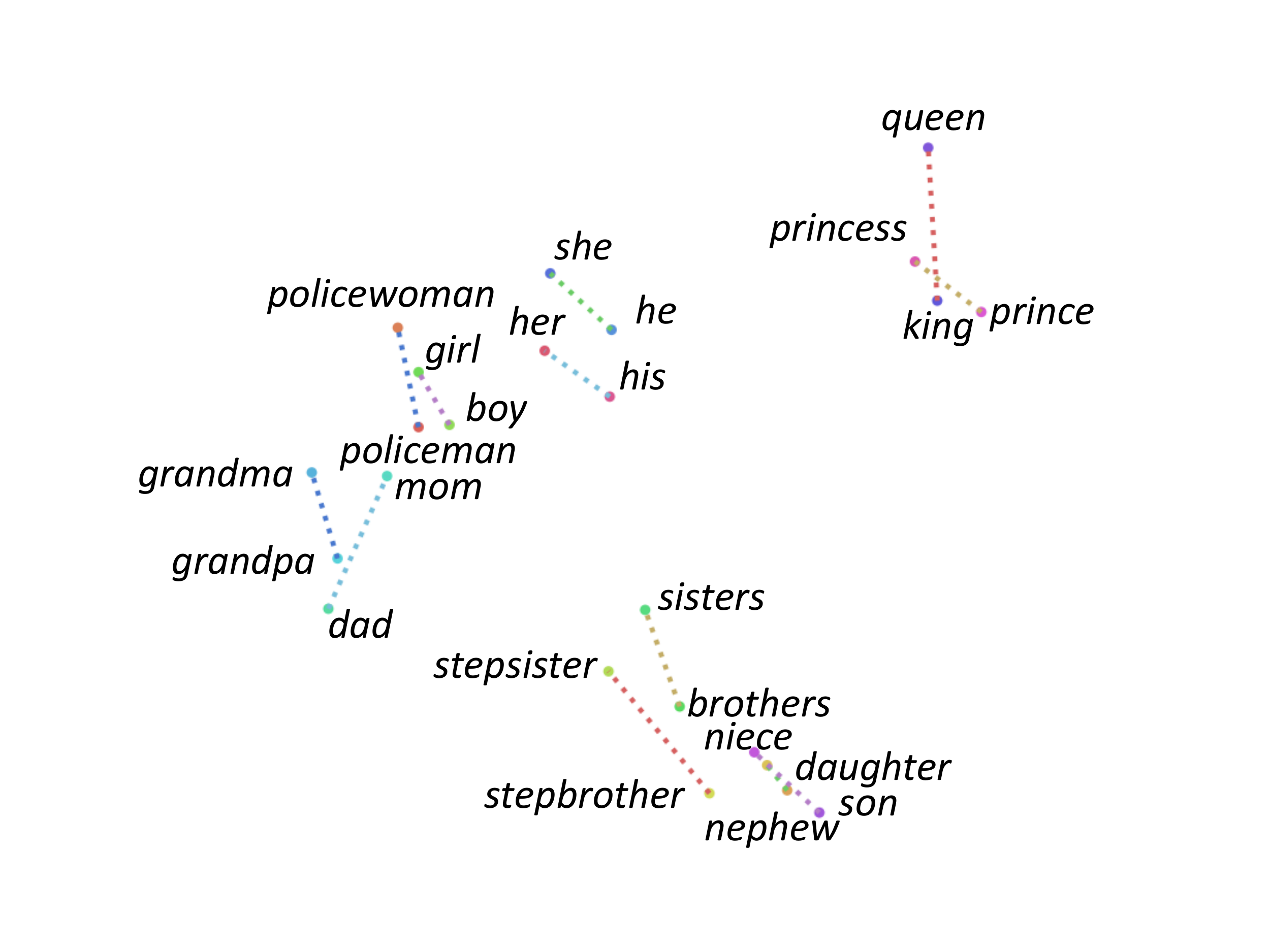}}
    \end{minipage}%
    }
	\subfigure[SynGCN]{
	\begin{minipage}[t]{0.3\linewidth}
	    \centering  
    	\fbox{\includegraphics[width=0.88\textwidth,trim= 80 60 60 60]{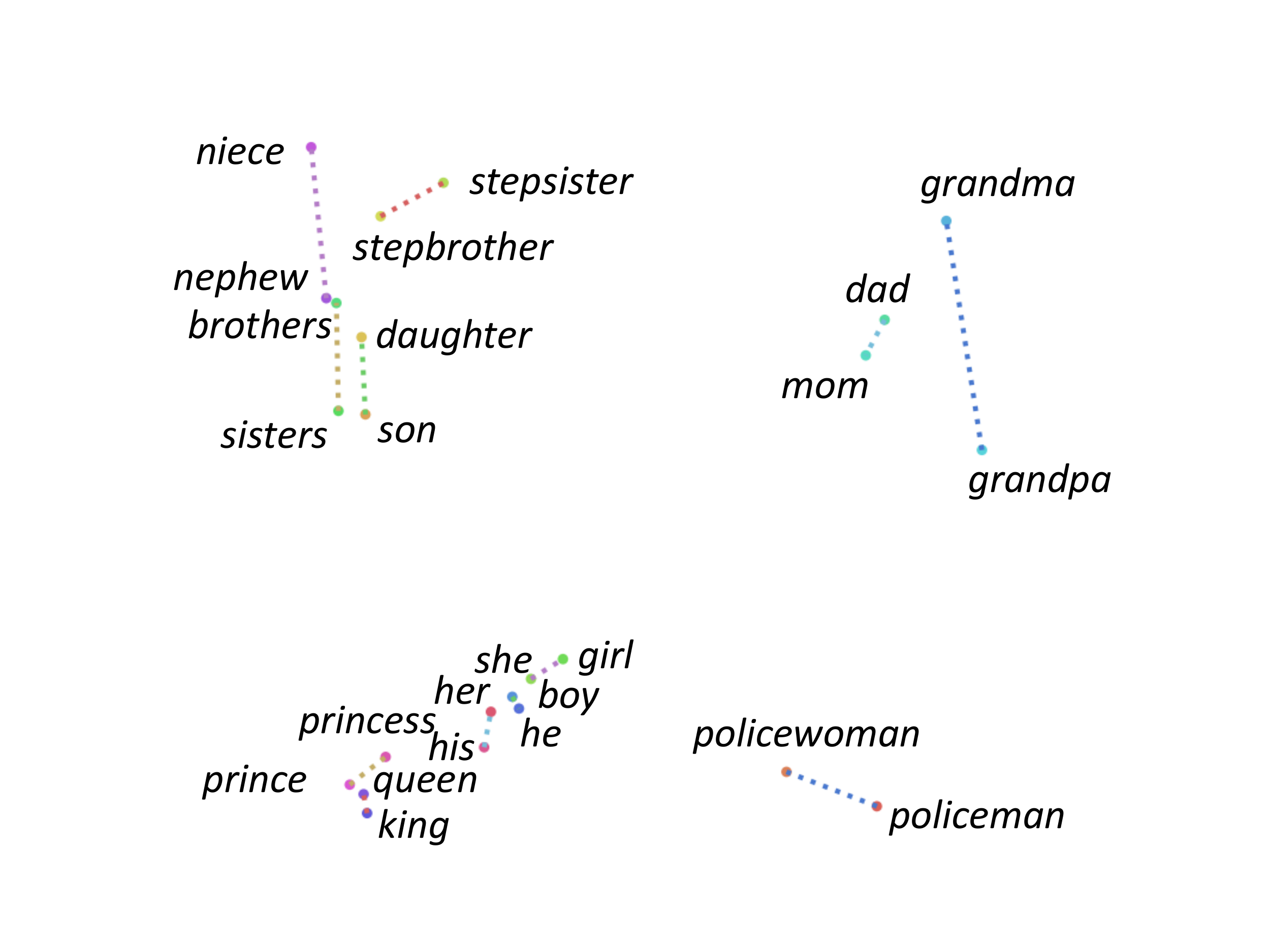}}
    \end{minipage}%
    }
    \subfigure[Ours]{
	\begin{minipage}[t]{0.3\linewidth}
	    \centering  
    	\fbox{\includegraphics[width=0.88\textwidth,trim= 80 60 60 60]{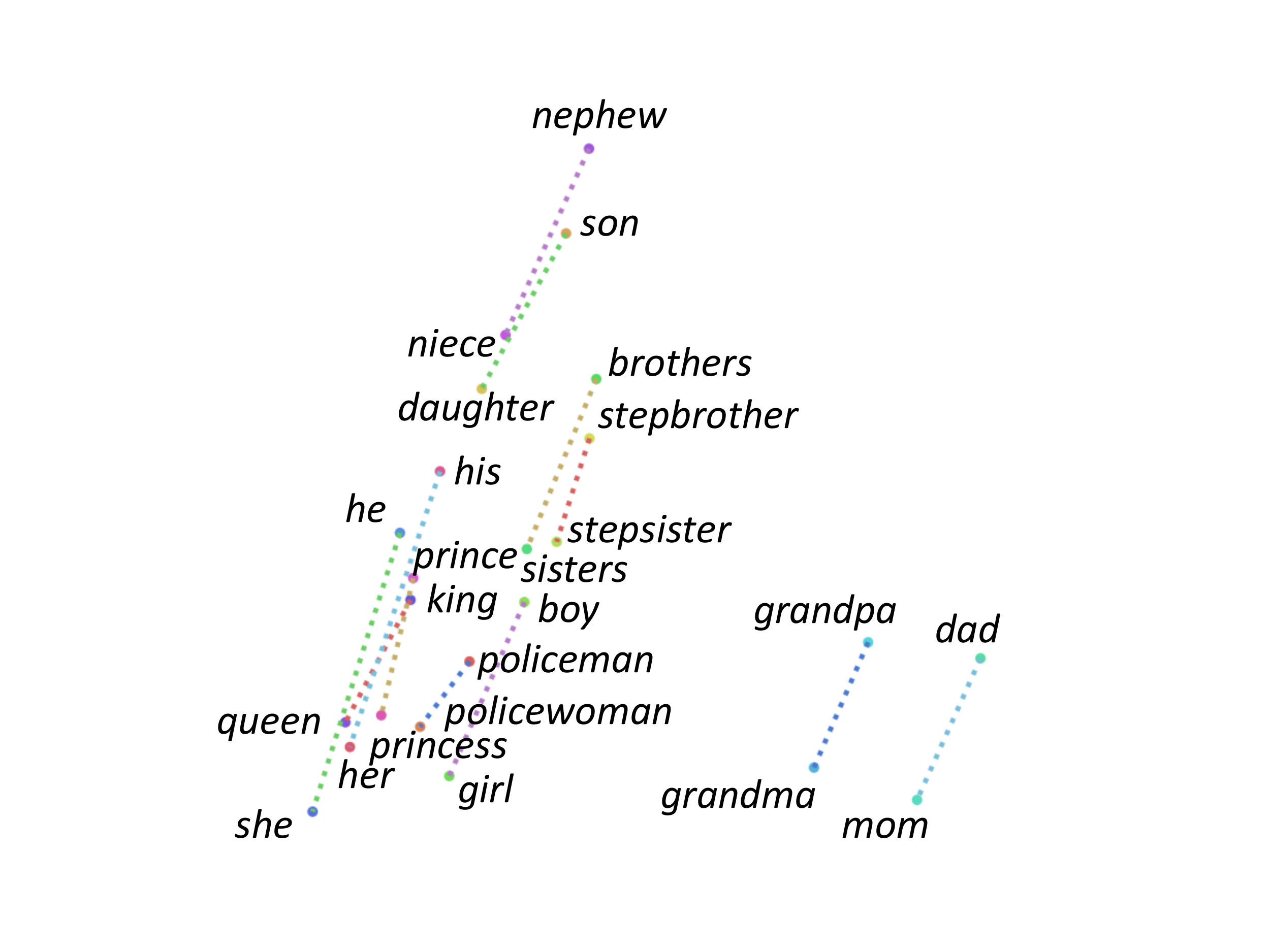}}
    \end{minipage}%
    }
    \centering
	\caption{Visualization of word pairs with the \textit{male-female} relationship.}
	\label{figure:tsne}
\end{figure*}
\section{Analysis}

Below we investigate the main reason behind the effectiveness of our method.

\noindent \textbf{Fine-grained Result.} Table~\ref{table:ana_ws} shows the word similarity results of some representative word pairs. BERT$_{token}$ does not capture the relatedness of $\langle$\textit{dividend}, \textit{payment}$\rangle$ and $\langle$\textit{murder}, \textit{manslaughter}$\rangle$ due to lack of consideration of context information, showing the discrepancy between human judgement and model scoring. BERT$_{avg}$ further improves the performance. Almost all the models give better results for word pairs that have higher co-occurrence probability. For example, the phrase ``\textit{benchmark index}'' and ``\textit{board recommendation}'' appear 8 and 29 times in corpus, respectively. In addition, the same neighboring words appearing in more sentences may have more similar averaged contextualized representations, thus leading BERT$_{avg}$ to give higher similarity scores compared with human judgement. SynGCN tends to underestimate the relationship between word pairs compared with other models, which shows negative influence of differentiating syntactic contexts. Overall, the results of our model are closer to human judgement.


\begin{table}[t]
	\centering
	\small
	\begin{tabular}{c|c}
		\hline 
		\textbf{Model}& \textbf{Nearest Neighbors for Word \textit{``while''}} \\
		\hline
		\multirow{2}{*}{SG}  &  \textit{whilst}, \textit{recuperating}, \textit{pursuing},  \\
		&   \textit{preparing}, \textit{attempting}, \textit{fending} \\
		\hline
		CBOW  &  \textit{whilst}, \textit{when}, \textit{still}, \textit{although}, \textit{and}, \textit{but} \\
		\hline
		GloVe  &  \textit{both}, \textit{taking}, \textit{`,'}, \textit{up}, \textit{but}, \textit{after} \\
		\hline
		FASTTEXT  &  \textit{whilst}, \textit{still}, \textit{and}, \textit{meanwhile}, \textit{instead}, \textit{though} \\
		\hline
		SynGCN  &  \textit{whilst}, \textit{time}, \textit{when}, \textit{years}, \textit{months}, \textit{tenures} \\ 
		\hline
		\multirow{2}{*}{Ours}  &  \textit{whilst}, \textit{whereas}, \textit{although},  \\
		 & \textit{conversely}, \textit{though}, \textit{meanwhile} \\
		\hline
	\end{tabular}
	\caption{Nearest neighbors for word \textit{``while''}.}
	\label{table:case}
\end{table}

For word analogy, we compare the performances of models according to different types of word pairs. Table~\ref{table:ana_google} shows the results. BERT$_{token}$ performs relatively lower on ``\textit{capital-country}" and ``\textit{city-state}" compared to skip-gram because it does not model context information. BERT$_{avg}$ improves the results by a large margin, giving comparable results on grammatical related word analogy such as ``\textit{plural}" due to the use of sentential information. SynGCN performs relatively well on grammatical related word pairs by using syntax structures. However, it does not perform well on ``\textit{capital-country}" and ``\textit{nationality-adjective}" word pairs compared with the sequential context based skip-gram model. In contrast, our model takes the advantages of these methods and gives the best overall performance.

\noindent \textbf{Nearest Neighbors.} Table~\ref{table:case} shows the nearest neighbors to the word \textit{``while''} according to cosine similarity. In particular, traditional methods yield words that tend to co-occur with the word ``\textit{while}'', such as ``\textit{preparing}'', ``\textit{still}'', ``\textit{taking}'' and ``\textit{instead}''. In contrast, SynGCN returns words that are semantically similar, namely those that are related to time. In contrast with the baselines, our method returns multiple conjunctions that have similar meanings to ``\textit{while}'', such as ``\textit{whilst}'', ``\textit{whereas}'' and ``\textit{although}'', which better conforms to the intuition, demonstrating the advantage of using dynamic embeddings to address word sense ambiguities.


\noindent \textbf{Word Pairs Visualization.} Figure~\ref{figure:tsne} shows the t-SNE~\cite{tsne} visualization results for word pairs with the {\it male-female} relationship. For example, the pronoun pair $\langle$\textit{he}, \textit{she}$\rangle$, the occupation pair $\langle$\textit{policeman}, \textit{policewoman}$\rangle$ and the family relation pair $\langle$\textit{grandpa}, \textit{grandma}$\rangle$  all differ only by the gender characteristics. Skip-gram, CBOW, GloVe, FASTTEXT and SynGCN baselines all capture the gender analogy through vector space topology to some extent. However, inconsistency exists between different word pairs. In contrast, the outputs of our method are highly consistent, better demonstrating the algebraic motivation behind skip-gram embeddings compared with the fully-static skip-gram algorithm. This demonstrates the effect of dynamic embeddings in better representing semantic information.

\begin{figure}[t]
	\centering  
	\includegraphics[scale=0.151]{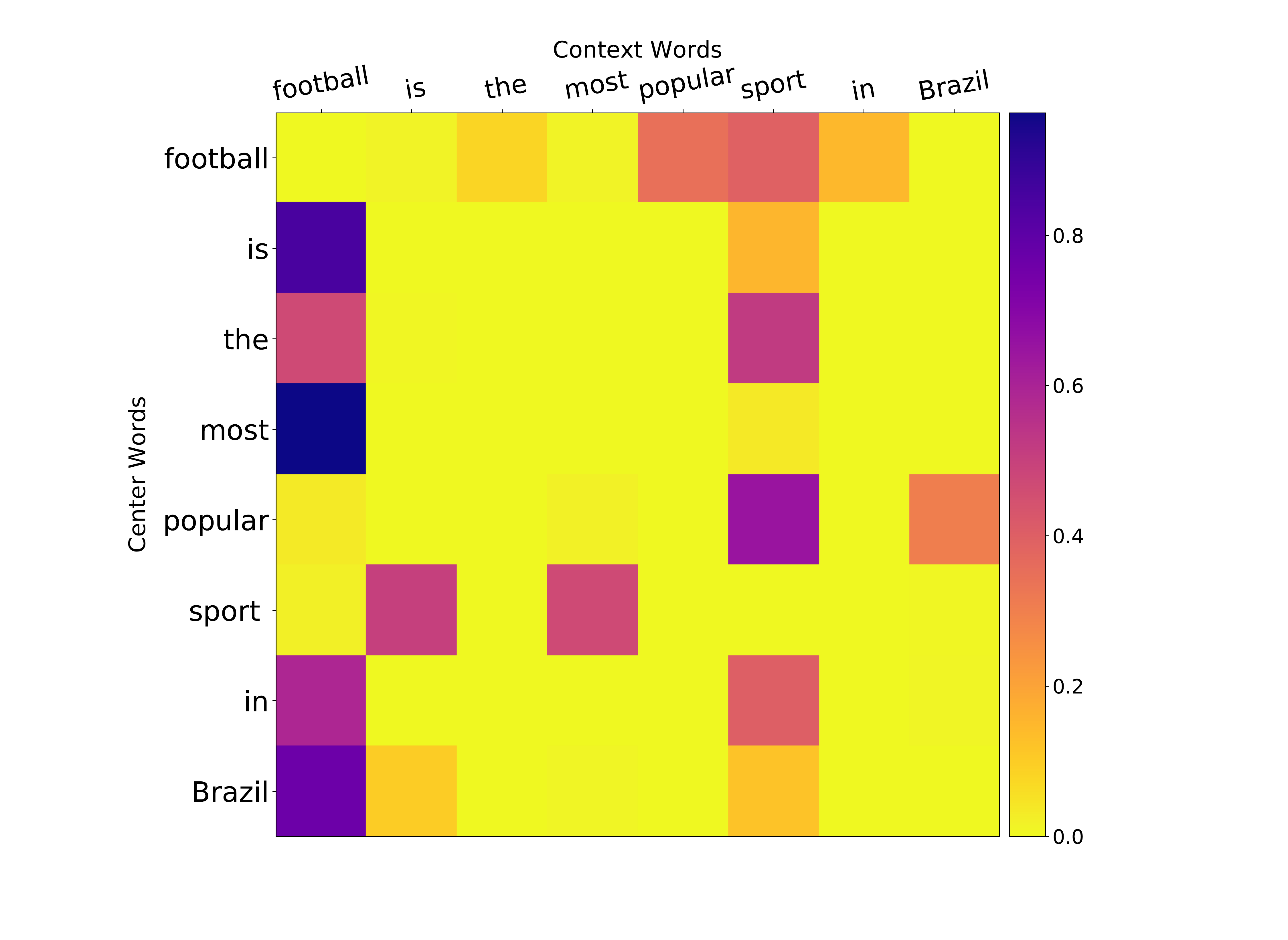}
	\caption{Attention distribution visualization of the sentence \textit{``football is the most popular sport in Brazil''}.}
	\label{figure:att}
\end{figure}

\noindent \textbf{Attention Distribution Visualization.} Figure~\ref{figure:att} shows the attention weights in Eq.~\ref{eq:att} when different words are used as the center word for the sentence ``\textit{football is the most popular sport in Brazil}''. As expected, for each center word, the most relevant context word receives relatively more attention. For example, the word ``\textit{football}'' is more associated with the words ``\textit{popular}'' and ``\textit{sport}'', and the word ``\textit{the}'' is more associated with nouns. No word pays attention to the word ``\textit{the}'' in the context words, which is a stop word.

\section{Conclusion}
We investigated how to make use of dynamic embedding for lexical semantics tasks such as word similarity and analogy, proposing a method to integrate dynamic embeddings into the training of static embeddings. Our method gives the best results on a range of benchmarks. Future work includes the investigation of sense embeddings and syntactic embeddings under our framework.
\bibliography{anthology,acl2020}
\bibliographystyle{acl_natbib}

\end{document}